\documentclass{article}
\usepackage{spconf,amsmath,epsfig}
\usepackage{xcolor}
\usepackage[super]{nth}
\usepackage{algorithm}
\usepackage{algcompatible}
\usepackage{graphicx}
\usepackage{epstopdf}
\usepackage{url}
\usepackage{multirow}
\newcommand{\GJP}[1]{\textcolor{black}{#1}}
\usepackage{subfigure}

\graphicspath{{eps/}}
\usepackage{atbegshi}
\AtBeginDocument{\AtBeginShipoutNext{\AtBeginShipoutDiscard}}

\let\OLDthebibliography\thebibliography
\renewcommand\thebibliography[1]{
  \OLDthebibliography{#1}
  \setlength{\parskip}{0pt}
  \setlength{\itemsep}{0pt plus 0.3ex}
}

\begin{document}\sloppy


\title{Deep Dictionary Learning with An Intra-class Constraint}
%

\name{Xia Yuan$^{1}$, Jianping Gou$^{1}$, Baosheng Yu$^{2}$, Jiali Yu$^{3}$ and Zhang Yi$^{4}$}
\address{$^{1}$School of Computer Science and Communication Engineering, Jiangsu University, Zhenjiang, China\\$^{2}$School of Computer Science, The University of Sydney, Darlington, NSW, Australia\\$^{3}$School of Mathematical Sciences, University of Electronic Science and Technology of China, China\\$^{4}$School of Computer Science, Sichuan University, Chengdu, China\\
 2212008046@stmail.ujs.edu.cn, goujianping@ujs.edu.cn, baosheng.yu@sydney.edu.au, \\yujiali@uestc.edu.cn, zhangyi@scu.edu.cn
}

\thanks{This work was supported in part by National Natural Science
Foundation of China (Grant No.61976107) and Qing Lan Project of Colleges and Universities of Jiangsu Province in 2020. }

\maketitle

\begin{abstract}
In recent years, deep dictionary learning (DDL)has attracted a great amount of attention due to its effectiveness for representation learning and visual recognition.~However, most existing methods focus on unsupervised deep dictionary learning, failing to further explore the category information.~To make full use of the category information of different samples, we propose a novel deep dictionary learning model with an intra-class constraint (DDLIC) for visual classification. Specifically, we design the intra-class compactness constraint on the intermediate representation at different levels to encourage the intra-class representations to be closer to each other, and eventually the learned representation becomes more discriminative.~Unlike the traditional DDL methods, during the classification stage, our DDLIC performs a layer-wise greedy optimization in a similar way to the training stage. Experimental results on four image datasets show that our method is superior to the state-of-the-art methods.
\end{abstract}
\begin{keywords}
Dictionary learning, Intra-class constraint, Image classification
\end{keywords}
\section{Introduction}
\label{1}
Dictionary learning has been widely used in various domains such as image denoising~\cite{kuruguntla2021study}, target detection~\cite{wang2021automatic}, clustering~\cite{fan2021robust}, and visual
classification~\cite{zhang2021geometry}. The motivation of dictionary learning
is to find a set of basis (also called dictionary atoms) from given samples,
and each sample can then be represented using the set of basis.~With the learned dictionary, we
can well approximate each sample using a linear combination of dictionary atoms and representations.

Sparse representation-based classification (or SRC) usually takes the training samples as the dictionary,
and then carries out sparse coding~\cite{wright2008robust}. However, using the original samples as a dictionary is problematic.~On the one hand, the dictionary is too large and the computational complexity is \GJP{too high} for large-scale classification problems. On the other hand, the original input training samples may contain noise, which leads to the inappropriate dictionary and suffers from the problem of poor robustness.~To address the above-mentioned issues, several supervised dictionary learning algorithms such as D-KSVD~\cite{zhang2010discriminative} and LC-KSVD~\cite{jiang2013label} have been proposed by \GJP{introducing category information} for dictionary learning. Meanwhile, many studies impose intra- or inter-class constraints on either representations or dictionaries on the basis
of traditional dictionary learning to obtain more discriminative representations or dictionaries.~For example, Foroughi et al.~\cite{foroughi2017object} proposed to build class-specific dictionaries to capture intra-class changes for datasets with significant intra-class changes.~Zeng et al.~\cite{zeng2020elm} used maximum edge criterion to make representations in low-dimensional space more discriminative. Inspired by SRC, Zhang et al.~\cite{zhang2021optimal} also suggested that the learned sparse representation should have a larger inter-class scatter and a smaller intra-class scatter.

With the rapid development of deep neural networks, people tend to explore the deep structure in different applications, including dictionary learning. Specifically, Tariyal et al.~\cite{tariyal2016deep} proposed a deep dictionary learning algorithm, that is, applying the shallow dictionary learning algorithm to the deep structure, and obtaining multiple dictionaries and corresponding representations in a greedy layer-by-layer fashion. Tang et al.~\cite{tang2020dictionary} presented a new deep dictionary learning and coding network (or DDLCN) for classification.~DDLCN replaces the convolutional layer in deep learning with dictionary learning and coding layer to obtain a more informative and discriminative representation at the final layer. \GJP{Singhai and Majumdar} ~\cite{singhal2020domain} applied the shallow coupled dictionary learning model to the deep structure, and solved the domain adaptive problem with the deep coupled dictionary learning, thus saving the computational cost. Recently, \GJP{Goel and Majumdar} ~\cite{goel2021sparse} highlighted the applicability of introducing a sparse subspace clustering loss term into DDL for hyperspectral image classification.

Deep dictionary learning is promising since its wide range of real-world applications, and we aim to learn multi-layer dictionaries \GJP{as well as deep discriminative} representations.~However, existing deep dictionary learning algorithms take into account the global constraints on representations but ignore the representation-based class information.~Inspired by this, we propose a novel deep dictionary learning model via designing an intra-class
constraint~(DDLIC) for visual classification.~Specifically, our model takes into account the category/class information in a multi-layer structure. In the proposed DDLIC model, we impose
a new intra-class constraint on the representation of each layer.~Since dictionary learning using a multi-layer architecture can also be regarded as \GJP{dimensional-reduction learning technology}, our
algorithm can also maintain the compactness of inter-class representation when projecting the high-dimensional data from original space to \GJP{low-dimensional} data in a low-dimensional space. Therefore, the representations obtained from the embedded space are more discriminative and more conducive for visual classification.  At the same time, in the classification stage of DDLIC, we use a similar way to the training stage. That is, for the original input test samples and the multi-layer dictionary obtained in the training stage, we can easily obtain the final representation in a layer-wise optimization manner.

\section{RELATED WORK}
\label{2}
In this section, we briefly review dictionary learning, including deep dictionary learning. \GJP{For convenient elaboration}, we use the notations defined as follows. Let the original input data from $C$ different classes, is $Z_0\in R^{k_0\times Cn}$, $k_0$ is the dimension of samples, $n_c$ is the number of training samples of class $c$,
where each class has the same training sample size $n$.
Given the deep learning model with $L$ layers. Let $D_l\in R^{k_{l-1}\times k_l}$, $(l=1,\ldots,L)$ denote the dictionary at the $l^{th}$ layer, where $k_l$ represents the number of atoms. $Z_l\in R^{k_l\times {Cn}}$ is the corresponding representation at the $l^{th}$ layer, $(l=1,\ldots,L)$. We further rewrite $Z_l$ as $Z_l=[Z_l^1,\cdots,Z_l^c,\cdots,Z_l^C]$, where $Z_l^c=[z_{l,1}^c,\cdots,z_{l,j}^c,\cdots,z_{l,n_c}^c]\in R^{k_l\times n_c}$. Here, $\varphi(\cdot)$ represents the activativation function. We have the test sample matrix $Y\in R^{k_0\times m}$, where $m$ is the test sample size.~During the classification stage, the corresponding representation at the $l^{th}$ layer can be denoted as $Z_{tl}~(l=1,\ldots,L)$.

\subsection{Dictionary Learning}
\label{2.1}
For the original input data $Z_0$, dictionary learning seeks to acquire a dictionary to represent the decomposition of data, which is composed of dictionary atoms. The objective of dictionary learning is formulated as follows \GJP{\cite{wright2008robust}}:
\begin{equation}\label{eq1}
\begin{aligned}
\min\limits_{D_1,Z_1}\|Z_0-D_1Z_1\|_F^2+\lambda \|Z_1\|_1,\\
s.t. \|d_{1,i}\|_2=1,~~\forall i\in[1,k_1]
\end{aligned}
\end{equation}
where $D_1=[d_{1,1},\ldots,d_{1,i},\ldots,d_{1,k_1}]$ and $\lambda$ is a parameter. To solve Eq.~(\ref{eq1}), many methods have appeared. A common strategy is to divide Eq.~(\ref{eq1}) into two steps, sparse coding and dictionary update, which are updated alternately until convergence. However, the discrimination of representation learned through shallow dictionaries is limited, so the classification results achieved may not be very precise.

\subsection{Deep Dictionary Learning}
\label{2.2}
In terms of classification tasks, deep dictionary learning achieves significant classification results and outperforms shallow dictionary learning.
The target objective function of DDL is as follows \GJP{\cite{tariyal2016deep}}:
\begin{equation}\label{eq2}
\min\limits_{D_1,\cdots,D_L,Z_L}\|Z_0-D_1\varphi(D_2\varphi(\cdots\varphi(D_LZ_L)))\|_F^2+\lambda \|Z_L\|_1.
\end{equation}
By introducing the identity activation function which means $\varphi(x)=x$, Eq.~(\ref{eq2}) can be converted into the following form:
\begin{equation}\label{eq3}
\min\limits_{D_1,\cdots,D_L,Z_L}\|Z_0-D_1D_2\cdots D_LZ_L\|_F^2+\lambda \|Z_L\|_1.
\end{equation}
We can obtain $Z_{l-1}=\varphi(D_lZ_l)$, $(l=1,\ldots,L)$, via defining $Z_l=\varphi(\cdots\varphi(D_LZ_L))$, and then we can easily figure out $\varphi^{-1}(Z_{l-1})=D_lZ_l$.
Till the penultimate layer, the representation of the different layers is dense in which we can easily solve it by the least square method. The optimization at the $l^{th}$ layer can be expressed as:
\begin{equation}\label{eq4}
\min\limits_{D_l,Z_l}\|Z_{l-1}-D_lZ_l\|_F^2.
\end{equation}
\begin{figure*}[!t]
\centering
\includegraphics[scale=0.94]{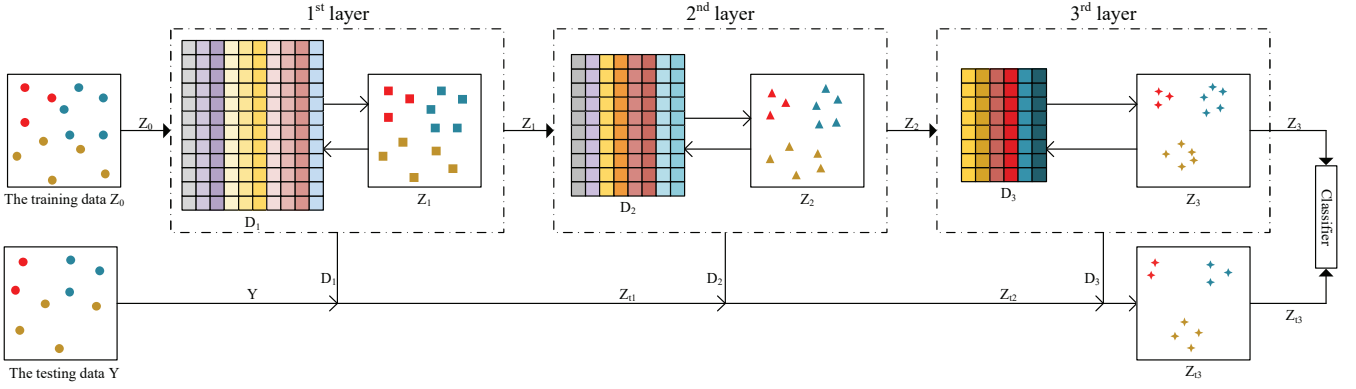}
\caption{The overview diagram of the three-level DDLIC model. \GJP{In the training phase, $Z_0$ is the training data, $D_l$~($l$=1,2,3) means the dictionary learned at the $l^{th}$ layer, and $Z_l$ ($l$=1,2,3) is the corresponding representation learned at the $l^{th}$ layer. In the classification phase, $Y$ is the test data, and $Z_{tl}$~($l$=1,2,3) is the representation based on the test sample obtained at the $l^{th}$ layer.}}\label{fig1}
\end{figure*}
For the last layer, we apply sparse constraints on the deepest level representation. In this way, the deepest level dictionary and its corresponding representation can be solved by the following objective function:
\begin{equation}\label{eq5}
\min\limits_{D_L,Z_L}\|Z_{L-1}-D_LZ_L\|_F^2+\lambda \|Z_L\|_1.
\end{equation}
It is obvious that Eq.~(\ref{eq5}) is a $l_1$ minimization problem, which can be easily solved with Iteration Soft Threading Algorithm.
During the classification stage, we first compute
\begin{equation}\label{eq6}
Z_{tL}=\mathop{\arg\min}_{Z_{tL}} \ \ \|Y-(D_1\cdots D_L)Z_{tL}\|_F^2+\lambda \|Z_{tL}\|_1.
\end{equation}
Finally, $Z_L$ and $Z_{tL}$ are the inputs to the KNN classifier, and thus the recognition accuracy of the image dataset is obtained.

\section{THE PROPOSED METHOD}\label{3}
In this section, we first introduce the motivation and then propose DDLIC algorithm with its optimization for visual classification.

\subsection{The Motivation}\label{3.1}

Existing deep dictionary learning methods mainly focus on learning global representations, but ignore the class information of representations at different layers.~Our main contribution is to introduce a novel intra-class constraint~\cite{yu2018correcting,yu2019deep} into the content of multi-layer dictionary learning.~At the same time, deep dictionary learning is equivalent to projecting the high-dimensional representation $Z_0$ of the original space into the low-dimensional representation $Z_L$ of the low-dimensional representation space.~Our goal is to find a proper low-dimensional representation space, where intra-class representations are in the same cluster, and there are obvious boundaries between different clusters.~In this way,~the final representation in the low-dimensional representation space is more discriminative and more conducive to visual classification.
In the DDLIC classification stage, we use a greedy learning method to learn the final representation of test samples layer by layer. This greedy learning method makes the relationship between layers closer in the classification stage, so as to obtain a better $Z_{tL}$ for the classification task. In Fig.~\ref{fig1}, we show a three-level DDLIC model as an intuitive example.

\subsection{The DDLIC Model}\label{3.2}

For deep dictionary learning, our model fully takes into account the compactness of intra-class representations in the process of multi-layer dictionary learning.
The overall DDLIC model can be formulated as:
\begin{equation}\label{eq7}
\begin{aligned}
\min_{D_1,\cdots,D_{L},Z_{L}}\sum_{l=1}^L(\sum_{c=1}^{C}\sum_{p=1}^{n_{c}}\|z_{l-1,p}^{c}-D_{l}z_{l, p}^{c}\|_{2}^{2}\\
+\alpha_l\sum_{c = 1}^{C}\sum_{i,j=1}^{n_{c}}\|z_{l,i}^{c}-z_{l,j}^{c}\|_{2}^{2}),
\end{aligned}
\end{equation}
where $\alpha_l$ is the regularization parameter of layer $l$,~$(l=1,\ldots,L)$. Specifically, the DDLIC objective function is composed of two terms as follows.~The first item is called the reconstruction loss item. It is reasonable to minimize the loss between different layers of the multi-layer dictionary learning structure.~One of our main contributions is to consider the relationship between intra-class representations in the multi-layer structure, that is, the second term in Eq.~(\ref{eq7}), which can be called the intra-class constraint term. The second term is proposed to increase the compactness of the intra-class representation at different layers. We can obtain dictionaries and representations at different levels through alternate optimization iterations.

\subsection{The DDLIC Training}\label{3.3}

Though the DDLIC model has $L$ different layers, the optimization process of each layer is similar. Here we take the solution to the $l^{th}$ layer of the DDLIC model as an example. The optimization objective function of the $l^{th}$ layer is defined as follows:
\begin{equation}\label{eq8}
\min_{D_{l},Z_{l}}\sum_{c=1}^{C}\sum_{p=1}^{n_{c}}\left\|z_{l-1, p}^{c}-D_{l}z_{l, p}^{c}\right\|_{2}^{2}+\alpha_l\sum_{c=1}^{C}\sum_{i,j=1}^{n_{c}}\left\|z_{l,i}^{c}-z_{l,j}^{c}\right\|_{2}^{2}
\end{equation}
By introducing
\begin{equation}\label{eq9}
F=\sum_{c=1}^{C}\sum_{p=1}^{n_{c}}\left\|z_{l-1, p}^{c}-D_{l}z_{l, p}^{c}\right\|_{2}^{2}+\alpha_l\sum_{c=1}^{C}\sum_{i,j=1}^{n_{c}}\left\|z_{l,i}^{c}-z_{l,j}^{c}\right\|_{2}^{2},
\end{equation}
We further decompose $F$ into two terms \GJP{including}
\begin{equation}\label{eq10}
F_1=\sum_{c=1}^{C}\sum_{p=1}^{n_{c}}\left\|z_{l-1, p}^{c}-D_{l}z_{l, p}^{c}\right\|_{2}^{2}
\end{equation}
and
\begin{equation}\label{eq11}
F_2=\alpha_l\sum_{c=1}^{C}\sum_{i,j=1}^{n_{c}}\left\|z_{l,i}^{c}-z_{l,j}^{c}\right\|_{2}^{2}.
\end{equation}
The $l^{th}$ layer's dictionary $D_l$ update stage: Only the reconstruction error term $F_1$ contains the $l^{th}$ layer dictionary $D_l$, so we can quickly find the updated expression of $D_l$ by using
\begin{equation}\label{eq12}
\begin{aligned}
\frac{\partial F}{\partial D_l} =\frac{\partial F_1}{\partial D_l} =&\frac{\partial (\|Z_{l-1}-D_lZ_l\|_F^2)}{\partial D_l} \\
=&-2Z_{l-1} Z_{l}^{T}+2 D_{l} Z_{l} Z_{l}^{T}.
\end{aligned}
\end{equation}
Let $\frac{\partial F}{\partial D_l}=0$, we have
\begin{equation}\label{eq14}
D_{l}=Z_{l-1} Z_{l}^{T}\left(Z_{l} Z_{l}^{T}\right)^{-1}.
\end{equation}
The $l^{th}$ layer's representation $Z_l$ update stage: Our goal is to find the partial derivative of $Z_l$ in $F$ and obtain the updated expression about $Z_l$. To promote intra-class compactness, we need to calculate the difference between a representation of a specific class and the remainder of the specific class at the $l^{th}$ layer. Consequently, we first find the partial derivative of the component $z_{l,k}^c$, and then place these components in order. Finally, we can get the desired $Z_l$. The partial derivative of the component $z_{l,k}^c$ is calculated as follows:
\begin{equation}\label{eq15}
\frac{\partial F_{1}}{\partial z_{l, k}^{c}}=-2 D_{l}^{T} z_{l-1, k}^{c}+2 D_{l}^{T} D_{l} z_{l, k}^{c},
\end{equation}
\begin{equation}\label{eq16}
\begin{aligned}
\frac{\partial F_{2}}{\partial z_{l,k}^{c}}
&=\alpha_l\frac{\partial}{\partial z_{l,k}^{c}}\left(\sum_{i,j=1}^{n_{c}}\left\|z_{l,i}^{c}-z_{l, j}^{c}\right\|_{2}^{2}\right)\\
&=2\alpha_l\frac{\partial}{\partial z_{l, k}^{c}}\left(\sum_{i \neq k}^{n_{c}}\left\|z_{l,i}^{c}-z_{l, k}^{c}\right\|_{2}^{2}\right)\\
&=4\alpha_l\sum_{i \neq k}^{n_{c}}\left(z_{l,k}^{c}-z_{l,i}^{c}\right)\\
&=4\alpha_l\left[\left(n_{c}-1\right)z_{l, k}^{c}-\sum_{i \neq k}^{n_{c}}z_{l,i}^{c}\right].
\end{aligned}
\end{equation}
Let
$\frac{\partial F}{\partial z_{l,k}^{c}}=\frac{\partial F_1}{\partial z_{l,k}^{c}}+\frac{\partial F_2}{\partial z_{l,k}^{c}}=0$,
we have
\begin{equation}\label{eq18}
\begin{small}
\begin{aligned}
&-2 D_{l}^{T} z_{l-1, k}^{c}+2 D_{l}^{T} D_{l} z_{l, k}^{c}+4 \alpha_l\left[\left(n_{c}-1\right) z_{l, k}^{c}-\sum_{i \neq k}^{n_{c}} z_{l, i}^{c}\right]=0 \\
&-2 D_{l}^{T} z_{l-1, k}^{c}+2 D_{l}^{T} D_{l} z_{l, k}^{c}+4 \alpha_l\left(n_{c}-1\right) z_{l, k}^{c}-4 \alpha_l \sum_{i \neq k}^{n_{c}} z_{l, i}^{c}=0 \\
&z_{l, k}^{c}=\left[D_{l}^{T} D_{l}+2 \alpha_l\left(n_{c}-1\right) I\right]^{-1}\Big(D_{l}^{T} z_{l-1, k}^{c}+2 \alpha_l \sum_{i \neq k}^{n_{c}} z_{l, i}^{c}\Big),
\end{aligned}
\end{small}
\end{equation}
where $I$ is the identity matrix. After obtaining each $z_{l,k}^c$, we finally get $Z_l$ as follows
\begin{equation}\label{eq19}
Z_l=\left[z_{l,1}^{1}, \cdots, z_{l, n_{1}}^{1},
\cdots, z_{l,1}^{C}, \cdots, z_{l, n_{C}}^{C}\right].
\end{equation}

\subsection{The DDLIC Classification}\label{3.4}
Unlike the typical DDL classification stage, the DDLIC classification stage takes a greedy, layer-by-layer approach to learn the deepest representation.
During the DDLIC classification stage, the $l^{th}$ layer representation can be calculated by the following formula:
\begin{equation}\label{eq20}
Z_{tl}=\underset{Z_{tl}}{\operatorname{argmin}} \|\left.Z_{t(l-1)}-D_{l}Z_{tl}\right\|_{2}^{2},
\end{equation}
where $l\in\{1,\ldots,L\}$. When $l=1$, we have $Z_{t0}=Y$.
Finally, we input the final level representation $Z_L$ in the training stage and the last representation $Z_{tL}$ obtained in the classification stage into the classifier. The overall DDLIC model is summarized in Algorithm~\ref{alg1}.

\begin{algorithm}[htb]
\caption{The DDLIC Algorithm}\label{alg1}
\begin{small}
\begin{algorithmic}
\REQUIRE
  Training matrix $Z_0$, \GJP{testing matrix $Y$,}  the number of layers $L$, the number of iterations K, and parameters $\alpha_1$,\ldots, $\alpha_L$.
\ENSURE
  The final classification accuracy.
\STATE 1.Initialize $D_1$, $\cdots$, $D_L$.
\FOR{$l = 1:L$}
\FOR {$k=1:K$}
\STATE 2.Update the dictionary $D_l$ using Eq.~(\ref{eq14}).
\STATE 3.Update the representation $Z_l$ using Eq.~(\ref{eq18}) and Eq.~(\ref{eq19}).
\ENDFOR
\ENDFOR
\FOR{$l = 1:L$}
\STATE 4.Update the representation $Z_{tl}$ using Eq.~(\ref{eq20}).
\ENDFOR
\STATE 5. $Z_L$ and $Z_{tL}$ are used as the input of the classifier.
\end{algorithmic}
\end{small}
\end{algorithm}

\section{Experiments}\label{4}

In this section, we evaluate the proposed DDLIC model on four datasets, including 1) the FEI and PIE29 datasets for face recognition; 2) the Shell dataset for image classification; and 3) the
Fifteen Scene Category dataset for scene recognition.

\subsection{Implementation Details}

In this paper, we use $L=3$, i.e., the three-layer deep dictionary learning with intra-class constraint~(DDLIC).
The $1^{st}$ layer dictionary is initialized by QR decomposition of the training sample matrix, the $2^{nd}$ layer dictionary and the $3^{rd}$ layer dictionary are randomly initialized. There are three parameters $\alpha_1$, $\alpha_2$, $\alpha_3$, we choose the optimal values via a small grid search, i.e., [1e-5, 1e-4, 1e-3, 1e-2, 0.1, 0.2]. In Sections \ref{4.1} and \ref{4.2}, we use seven comparison algorithms, there are three shallow dictionary learning algorithms, namely SRC \cite{wright2008robust}, D-KSVD \cite{zhang2010discriminative} and LC-KSVD \cite{jiang2013label}, two deep learning methods, namely stacked autoencoder (SAE) \cite{bengio2009learning} and deep belief network (DBN) \cite{hinton2009deep}, and two deep dictionary learning algorithms, called DDL \cite{tariyal2016deep} and DDLCN \cite{tang2020dictionary}. In the classification stage, we choose KNN as the final classifier. Note that the neighbour size of the KNN classification is optimally determined in the range of 1 to 30 with step 1.~The number of iterations in each layer is fixed at 20. For each dataset, $h$ samples from each dataset were selected as training samples and the remainder as test ones. We randomly generated 10 training-test splits, and reported the average classification accuracy. It is noteworthy that we also tuned the running parameters of all the competitors.

\begin{table}[!h]
\centering
\caption{The accuracies~(\%) of the competing methods on the FEI and PIE29 datasets.}\label{tb1}
\begin{small}
\setlength{\tabcolsep}{1.5mm}{
\begin{tabular}{llll|lll}
\\\hline
Method  & \multicolumn{3}{c}{FEI} & \multicolumn{3}{c}{PIE29 } \\\hline
\emph{h}       & 6          & 8          & 10        & 13          & 15         & 17         \\\hline
SRC     & 84.34      & 89.83      & 93.75     & 92.81       & 93.59      & 94.37      \\
D-KSVD  & 69.19      & 70.49      & 70.69     & 93.69       & 94.41      & 94.62      \\
LC-KSVD & 70.75      & 71.52      & 72.32     & 93.76       & 93.90      & 94.56      \\\hline
DBN     & 80.06      & 84.75      & 85.12     & 93.62       & 93.92      & 94.33      \\
SAE     & 81.42      & 82.56      & 88.63     & 93.72       & 93.96      & 94.92      \\\hline
DDL     & 79.31      & 84.25      & 86.75     & 92.51       & 94.38      & 95.42      \\
DDLCN   & 80.63      & 81.93      & 82.57     & 93.93       & 94.35      & 95.04      \\\hline
\textbf{DDLIC}   & \textbf{87.81} & \textbf{93.33} & \textbf{95.87}  & \textbf{94.12}   & \textbf{94.65 } & \textbf{96.22  }    \\\hline
\end{tabular}}
\end{small}
\end{table}

\subsection{Face Recognition}\label{4.1}

The FEI dataset contains 2800 images for 200 individuals, each of which has 14 color images with the original size of 640x480 pixels. The PIE29 dataset is a subset of the CMU PIE dataset. The PIE29 dataset consists of 68 people, each of which has 24 face images including front face, left and right face, front face smile and illumination from different angles. In the experiments, we resize each face image to 28$\times$28 pixels and the dictionary atoms are 400, 200, and 100 in the three-layer DDLIC model, respectively. The classification accuracies of different competing methods on the FEI and PIE29 dataset are shown in Table~\ref{tb1}.
The experimental results show that the higher the number of training samples, the higher the classification accuracy. Compared with other comparison algorithms, our method consistently achieves higher classification accuracy.

\begin{table}[!h]
\centering
\caption{The classification accuracies~(\%) of the competing methods on the Shell and Fifteen Scene Category datasets.}\label{tb2}
\begin{small}
\setlength{\tabcolsep}{1.5mm}{
\begin{tabular}{llll|lll}
\\\hline
Method  & \multicolumn{3}{c}{Shell} & \multicolumn{3}{c}{Fifteen Scene Category} \\\hline
$h$       & 10          & 12         & 14         & 34              & 42              & 50              \\\hline
SRC     & 74.27       & 77.23      & 78.90      & 96.24           & 96.92           & 97.22           \\
D-KSVD  & 54.94       & 56.64      & 57.08      & 91.60           & 93.33           & 93.69           \\
LC-KSVD & 56.80       & 60.85      & 61.94      & 92.28           & 93.08           & 93.68           \\\hline
DBN     & 71.27       & 74.72      & 76.74      & 95.08           & 95.51           & 96.14                 \\
SAE     & 73.43       & 75.84      & 77.49      & 94.27           & 95.36           & 95.96            \\\hline
DDL     & 69.18       & 71.36      & 74.13      & 95.74           & 96.56           & 97.25                \\
DDLCN   & 63.68       & 71.82      & 72.22      & 95.41           & 95.70           & 96.82      \\\hline
\textbf{DDLIC}   & \textbf{74.61}       & \textbf{78.37}     & \textbf{79.11}      & \textbf{96.32}           & \textbf{97.22}           & \textbf{98.03}           \\\hline
\end{tabular}}
\end{small}
\end{table}

\subsection{Image Recognition}\label{4.2}

The Shell dataset includes 2680 shell images from 134 classes and each class contains 20 images, each of which was resized into 28$\times$28 pixels in the experiments.~The Fifteen Scene Category dataset includes fifteen natural scene categories, each of which contains 200-400 images.~For each image in the Fifteen Scene Category dataset, we extract a 1000-d spatial pyramid feature~\cite{lazebnik2006beyond} in our experiments.
For the Shell dataset, the dictionary atoms are 400, 200, and 100 in the three-layer DDLIC model, respectively. For the Fifteen Scene Category dataset, we choose dictionary atoms in the three-layer DDLIC model to be 500, 300, and 100, respectively.~Table~\ref{tb2} shows the comparison experiments on these two image datasets. We can see that our DDLIC consistently outperforms the competing methods. Thus, it can be concluded from these experimental results above that the proposed DDLIC can learn discriminative representation through deep dictionary learning with the designed intra-class constraint.

\begin{figure}[!ht]
\centering
  \subfigure[The original representation $Z_0$.]
  {\includegraphics[scale=0.287]{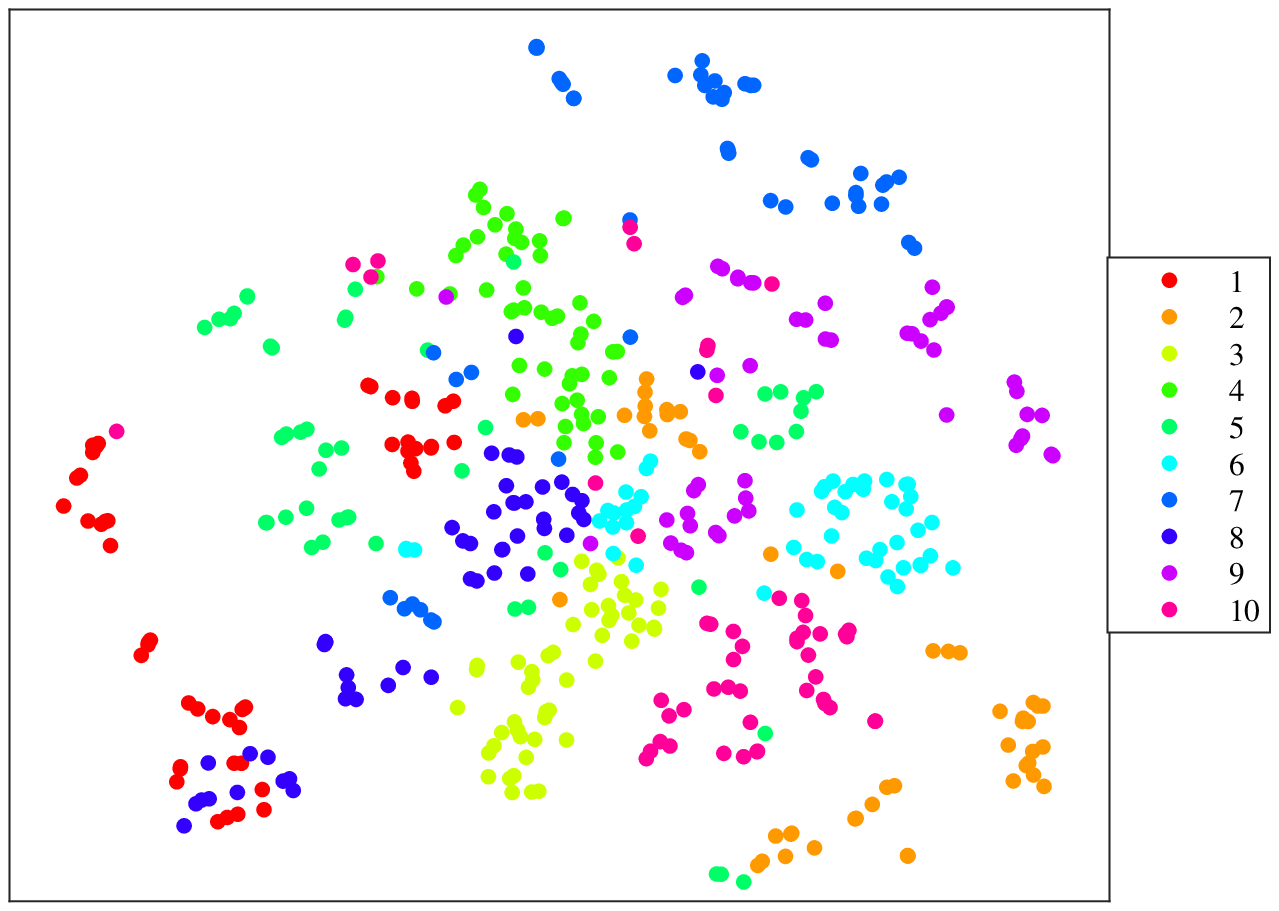}}
  \subfigure[The $1^{st}$ layer representation $Z_1$.]
  {\includegraphics[scale=0.287]{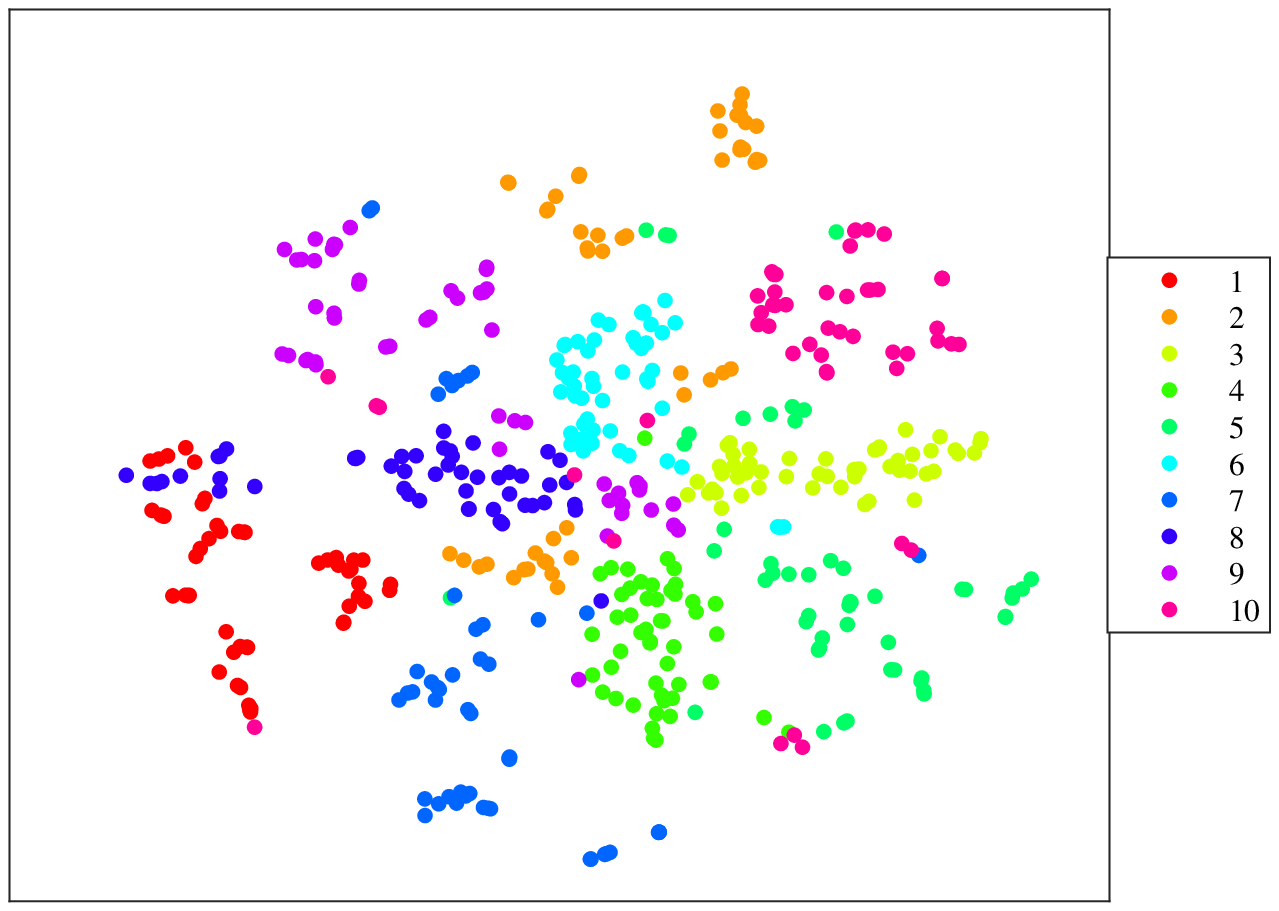}}
  \subfigure[The $2^{nd}$ layer representation $Z_2$.]
  {\includegraphics[scale=0.287]{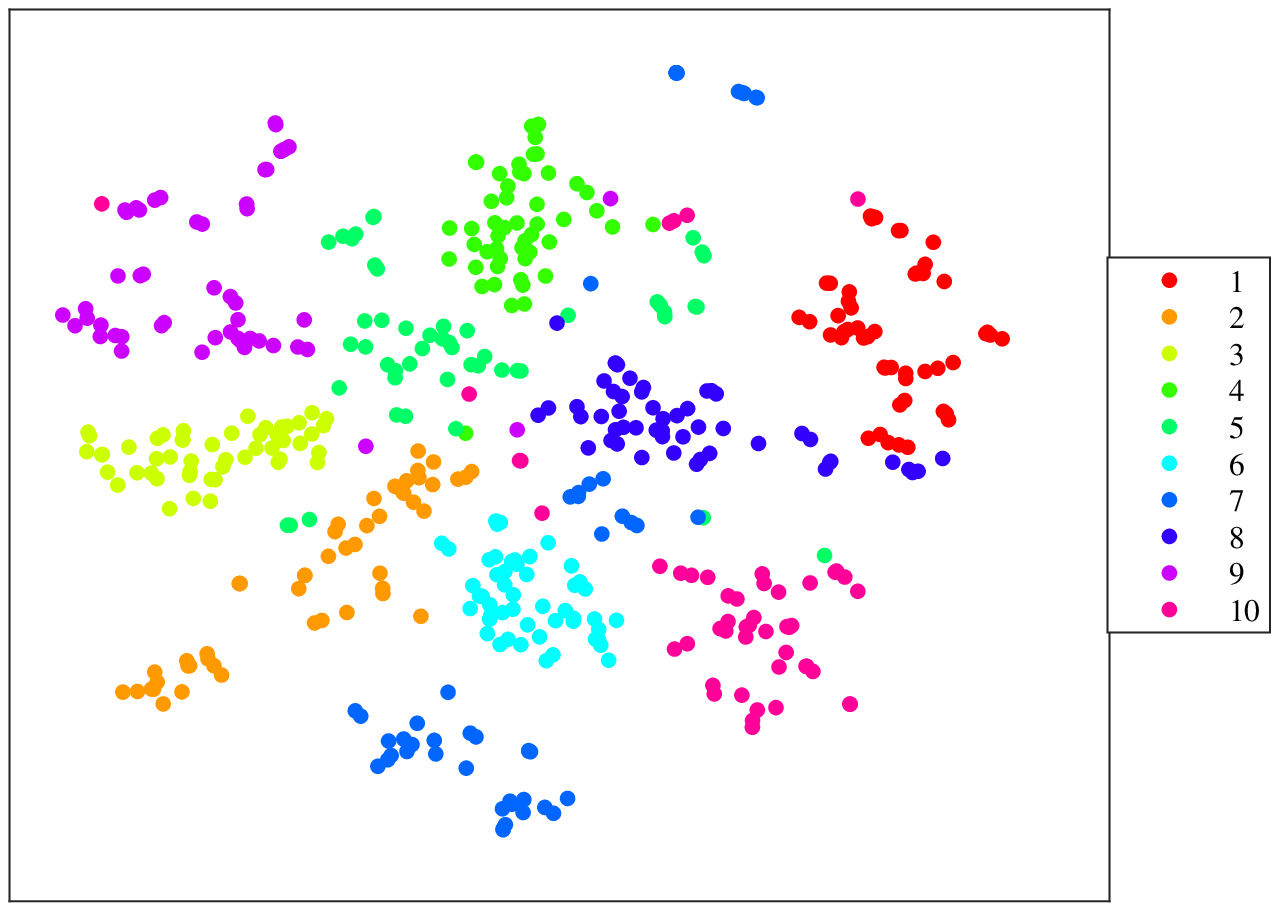}}
  \subfigure[The $3^{rd}$ layer representation $Z_3$]
  {\includegraphics[scale=0.287]{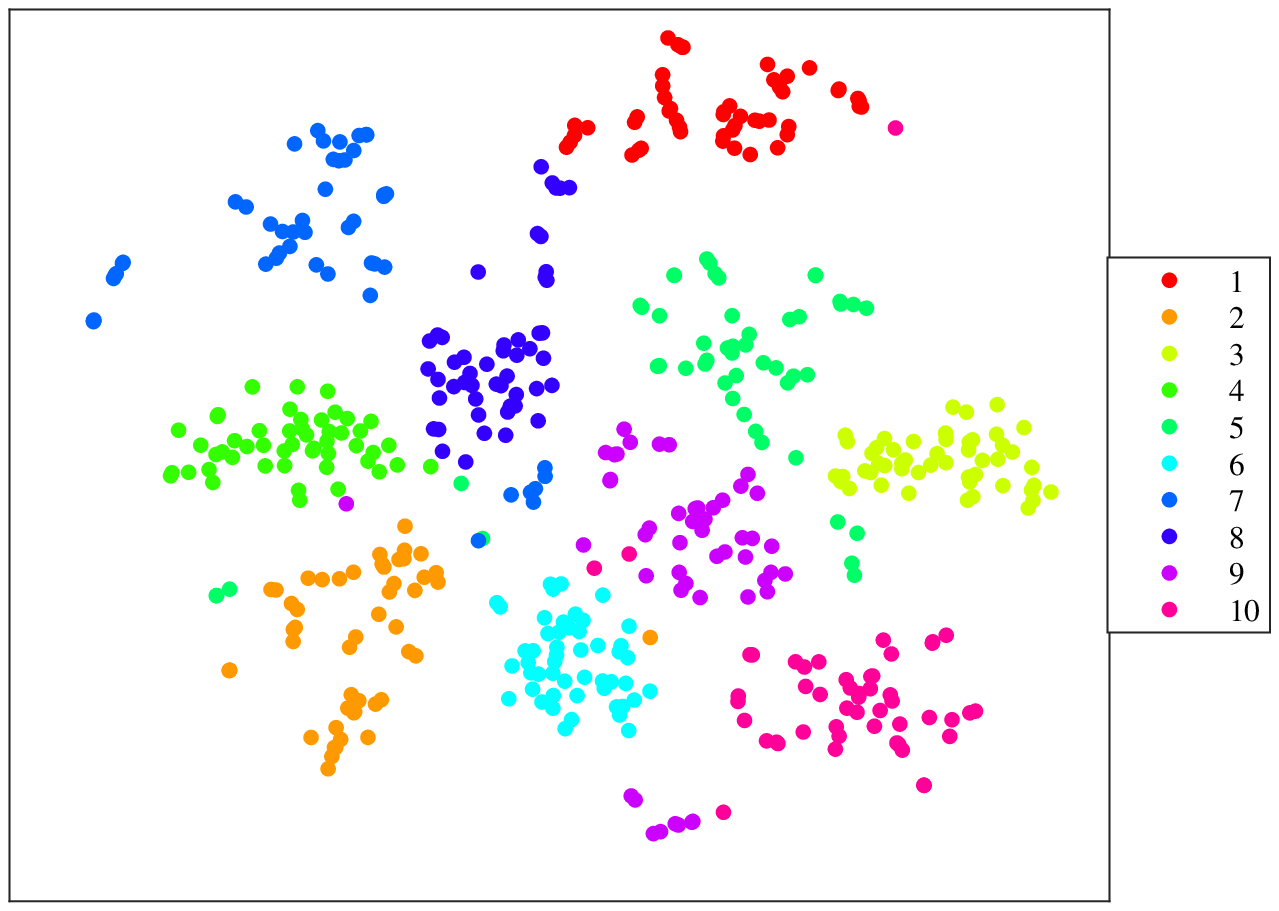}}
  \caption{The visualizations of the original input data $Z_0$, the $1^{st}$ layer representation $Z_1$, the $2^{nd}$ layer representation $Z_2$, and the $3^{rd}$ layer representation $Z_3$ on the Fifteen Scene Category.} \label{fig2}
\end{figure}

\begin{figure}[!ht]
\centering
\includegraphics[width=0.9\linewidth]{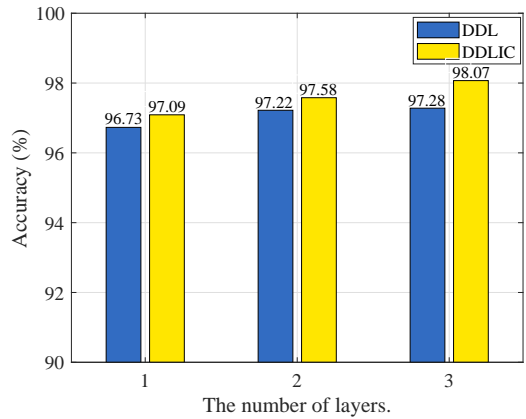}
\caption{The comparative classification results of the $1^{st}$, $2^{nd}$ and $3^{rd}$ layer representations on the Fifteen Scene Category dataset learned by DDL and DDLIC, respectively.}\label{fig3}
\end{figure}

\subsection{Hierarchical Discrimination Analysis}\label{4.3}

In this section, we will further explore the pattern discrimination of the original data $Z_0$ and the representations $Z_1$, $Z_2$, $Z_3$ learned in the three-layer DDLIC model.~We first take the Fifteen Scene Category dataset as an example and select 50 training samples for each class. Then, our DDLIC is performed on one training-test split. In these experiments, we illustrate the original data $Z_0$ and the representations $Z_1$, $Z_2$ and $Z_3$ from the randomly chosen 10 classes. The $Z_0$, $Z_1$, $Z_2$ and $Z_3$ are visualized in two-dimensional spaces by using t-SNE \cite{van2008visualizing}, shown in \GJP{Fig.~\ref{fig2}}.~It is clear \GJP{that} the original data points for $Z_0$ have poor discrimination among all the classes. By imposing our novel intra-class representation constraint on each layer, the intra-class representation becomes more compact with the deepening layer, so that the representation for deeper level becomes more discriminative. Meanwhile, \GJP{Fig.~\ref{fig3}} shows that the classification results using the $1^{st}$, $2^{nd}$ and $3^{rd}$ layer representations on the Fifteen Scene Category dataset learned by DDL and DDLIC, respectively. For DDL and DDLIC, as the number of dictionary layers increases, the deeper representations become more discriminative with the higher classification accuracy. More importantly, the layer-specific representation in our DDLIC is more discriminative with the higher classification accuracy than the corresponding one in DDL.

\vspace*{-2mm}

\section{Conclusion}\label{5}
In this paper, we propose a novel DDL-based method, named deep dictionary learning with an intra-class constraint for visual classification.~To explore the category information of representations, we introduce a new intra-class constraint term into the traditional DDL to make the intra-class representations more compact, so that in the multi-layer architecture, the final-layer representation is more discriminative, and can be used to improve the classification accuracy. Last but not least, in DDLIC, we take the idea of layer-by-layer learning to establish a new optimization model for the classification stage. Extensive experiments show the effectiveness of the proposed DDLIC model for different visual classification tasks. Since this paper only considers from the perspective of intra-class relations, deep dictionary learning with the representation-based intra-class and inter-class relationship constraints will be the subject of future study.

%

{\small
\bibliographystyle{IEEEbib}
\bibliography{DDLIC}
}
\end{document}